\pdfoutput=1
\documentclass[11pt]{article}

\usepackage[final]{acl}

\usepackage{times}
\usepackage{latexsym}
\usepackage{enumitem}  % 提供内联列表环境
\usepackage[T1]{fontenc}
\usepackage[utf8]{inputenc}

\usepackage{microtype}

\usepackage{inconsolata}

\usepackage{graphicx}

\usepackage{amsmath}
\usepackage{bm}
\usepackage{amssymb}
\usepackage{multirow}
\usepackage{booktabs}
\usepackage{amssymb}
\usepackage{color}

\title{Beyond Completion: A Foundation Model for General Knowledge Graph Reasoning}

\author{
  Yin Hua\textsuperscript{$\spadesuit$},
  Zhiqiang Liu\textsuperscript{$\spadesuit$},
  Mingyang Chen\textsuperscript{$\spadesuit$},
  Zheng Fang\textsuperscript{$\clubsuit$},
  Chi Man Wong\textsuperscript{$\clubsuit$ $\diamondsuit$}, \\
  \textbf{Lingxiao Li}\textsuperscript{$\clubsuit$},
  \textbf{Chi Man VONG}\textsuperscript{$\diamondsuit$},
  \textbf{Huajun Chen}\textsuperscript{$\spadesuit$ $\heartsuit$},
  \textbf{Wen Zhang}\textsuperscript{$\spadesuit$†}  % 在这里手动加 †
  \\
  $\spadesuit$ Zhejiang University \\
  $\clubsuit$ Shopee Pte.Ltd.,
  $\diamondsuit$ University of Macau \\
  $\heartsuit$ Zhejiang Key Laboratory of Big Data Intelligent Computing    \\
  \texttt{\{22351088,zhang.wen\}@zju.edu.cn}
}

\begin{document}
\maketitle

%——下面这几行把脚注样式切换到“符号”格式，并插入 † 通信作者脚注
\begingroup
  \renewcommand{\thefootnote}{\fnsymbol{footnote}}
  \setcounter{footnote}{0}
  \footnotetext{† Corresponding author}
\endgroup

\begin{abstract}
% In natural language processing (NLP) and computer vision (CV), successfully applying pre-trained models for cross-domain transfer has proven their significant potential. However, despite the rich structural and textual information in knowledge graphs (KGs), existing research has predominantly focused on graph structure, with most efforts confined to knowledge graph completion (KGC) tasks, hindering their cross-domain applicability. In this study, we introduce MERRY, a general knowledge graph reasoning framework, and investigate its capabilities in two cross-domain tasks: zero-shot link prediction and 
% knowledge graph question answering (KGQA)
% % \fz{KGQA (maybe need to explain the abbr?)}
% . Specifically, we propose a multi-perspective Conditional Message Passing (CMP) 
% % \cm{need long form for first appear}
% encoding architecture to bridge the gap between textual and structural modalities, enabling their effective integration. Furthermore, we introduce a dynamic residual fusion module that dynamically retains relevant portions of the original textual modality information and a flexible edge scoring mechanism that enables adaptation to a wide range of downstream tasks. Extensive evaluation across 25 datasets demonstrates that MERRY outperforms existing baselines in most cases, exhibiting strong generalization within the KG domain and excellent transferability to the KGQA cross-domain task.
In natural language processing (NLP) and computer vision (CV), the successful application of foundation models across diverse tasks has demonstrated their remarkable potential. However, despite the rich structural and textual information embedded in knowledge graphs (KGs), existing research of foundation model for KG has primarily focused on their structural aspects, with most efforts restricted to \textbf{in-KG} tasks (e.g., knowledge graph completion, KGC). This limitation has hindered progress in addressing more challenging \textbf{out-of-KG} tasks. 
% limiting broader applications. 
In this paper, we introduce MERRY, a foundation model for  general knowledge graph reasoning, and investigate its performance across two task categories: \textbf{in-KG} reasoning tasks (e.g., KGC) and \textbf{out-of-KG} tasks (e.g., KG question answering, KGQA). We not only utilize the structural information, but also the textual information in KGs. Specifically, we propose a multi-perspective Conditional Message Passing (CMP) encoding architecture to bridge the gap between textual and structural modalities, enabling their seamless integration. Additionally, we introduce a dynamic residual fusion module to selectively retain relevant textual information and a flexible edge scoring mechanism to adapt to diverse downstream tasks. Comprehensive evaluations on 28 datasets demonstrate that MERRY outperforms existing baselines in most scenarios, showcasing strong reasoning capabilities within KGs and excellent generalization to out-of-KG tasks such as KGQA.
\end{abstract}

\section{Introduction}

Knowledge graphs (KGs) are structured knowledge bases that represent entities and their relationships, providing a foundation for reasoning and information retrieval in various real-world domains. With their rich entity representations and rigorous logical connections, KGs have become integral to applications such as classification \cite{liu-etal-2023-enhancing_kg_cls}, recommendation \cite{guo2020surveyknowledgegraphbasedrecommender_kg_recommendation}, knowledge retrieval \cite{Xu_2024_kg_retriver}, and QA systems \cite{DBLP:journals/tnn/JiPCMY22}, as well as knowledge-grounded LLM alignment \cite{lzq}.

Recently, foundation  models in NLP and CV \citet{raffel2023exploringlimitstransferlearning, DBLP:journals/rfc/rfc9405, li2024llava, ravi2024sam2} have demonstrated significant advancements in transfer learning, enabling improved performance across datasets and tasks. Inspired by these successes, researchers have developed foundational models for KGs that aim to generalize across datasets and adapt to diverse reasoning tasks. 
KGs naturally encompass both structural and textual information, yet existing research has predominantly focused on leveraging their structural aspects, with relatively limited attention to the textual modality \cite{galkin2023ultra, zhu2021neural, Teru2020InductiveRP_grail, geng2022relationalmessagepassingfully_rmpi, MaKEr, liu2024unihr}. 
However, fully utilizing both modalities is crucial, as textual information provides contextual knowledge that complements structural representations. This integration is particularly important for downstream applications such as commonsense reasoning and KGQA, where the combination of relational and contextual knowledge significantly enhances task performance \cite{yasunaga2021qagnn, zhang2021greaselm, markowitz-etal-2022-statik}.
In addition, prior work has largely been restricted to in-KG reasoning tasks, such as KG Completion (KGC), and has not adequately addressed the challenges posed by out-of-KG reasoning tasks, such as KGQA. out-of-KG tasks require models to generalize beyond the explicit structure of KGs, incorporating both modalities to handle more complex reasoning scenarios. 

Overcoming these limitations involves addressing three key challenges in model design: (1) mitigating the semantic disparity between textual and structural information to facilitate effective integration; (2) balancing the contributions of textual and structural modalities to suit diverse task requirements, particularly for reasoning beyond KGs; and (3) maintaining an unbiased training procedure to enable robust generalization across datasets without favoring specific entities or relations \cite{wang-etal-2022-simkgc, markowitz-etal-2022-statik}.

To address these challenges, we propose the Multi-pErspective Reasoning sYstem, MERRY, a universal knowledge graph reasoning framework. MERRY integrates textual and structural information through a global structural semantic encoding module (GCMP), designed to reconcile their semantic differences. 
To enhance adaptability, we introduce a dynamic text-adaptive fusion module (DTAF) that selectively preserves essential textual information, facilitating effective application across a range of tasks. Furthermore, we develop a flexible edge scoring mechanism that adjusts adaptively to meet the specific requirements of downstream tasks, thereby enhancing the model's transferability across diverse reasoning scenarios.
% To enhance adaptability, we introduce a dynamic text-adaptive fusion module (DTAF) that selectively preserves essential textual information for a wide range of tasks. Additionally, a flexible edge scoring mechanism is designed to meet downstream task requirements, improving transferability across reasoning scenarios.
% To enhance adaptability, we introduce a dynamic residual fusion module (DTAF) that selectively retains relevant textual information, ensuring effective application across tasks. Finally, we design a flexible edge scoring mechanism that \fz{too many "dynamic", paraphrase it} dynamically adjusts to downstream task requirements, improving transferability across reasoning scenarios.

% \fz{Rewrite: Both within-KG (zero-shot KGC) and beyond-KG (KGQA) tasks are evaluated in our MERRY.} We evaluate MERRY on two representative tasks: zero-shot KGC (a within-KG reasoning task) and KGQA (a beyond-KG reasoning task). 
Both in-KG (zero-shot KGC) and out-of-KG (KGQA) tasks are evaluated in our MERRY.
Results across 28 datasets demonstrate that MERRY consistently outperforms multiple benchmark models in both tasks, highlighting its robust generalization and adaptability. Our codes are released to the GitHub\footnote{\url{https://github.com/zjukg/MERRY}}. The main contributions of this paper are as follows: 
\begin{itemize} 
\item We propose a novel framework for addressing in-KG and out-of-KG reasoning tasks, integrating textual and structural modalities.
% \item The MERRY framework incorporates multi-perspective CMP encoding, a dynamic residual fusion module, and a flexible edge scoring mechanism to address reasoning \fz{replace "challenges" with more specific description, like "bias", "conflict", "imbalance", etc.} challenges across diverse KG tasks. 
\item We propose MERRY as a foundation model for general KG reasoning.
By harmonizing structural and textual information, the framework achieves effective integration and ensures smooth transferability across reasoning tasks with varying modality demands.
\item We validate MERRY’s performance on 28 datasets, demonstrating its effectiveness in zero-shot KGC and KGQA, with consistent improvements over multiple benchmarks. \end{itemize}

\section{Related Work}

\paragraph{Inductive Knowledge Graph Completion}
KG Completion (KGC) is a fundamental task for reasoning over knowledge graphs. Its evolution can be categorized into three stages. Early work focused on the transductive setting, where KGs are static, and entity and relation representations are precomputed and stored \cite{bordes2013translating_transe, sun2019rotateknowledgegraphembedding_rotate, vashishth2020compositionbasedmultirelationalgraphconvolutional_compgcn}.

Real-world KGs, however, are dynamic \cite{cui2022inductiveknowledgegraphreasoning_argcn}, requiring inductive methods to handle unseen entities and relations \cite{Teru2020InductiveRP_grail, geng2022relationalmessagepassingfully_rmpi}. These approaches rely on supervised training, limiting their generalization to unseen datasets and diverse KGC tasks.

Recent efforts leverage pre-training paradigms from NLP and CV. For example, ULTRA \cite{galkin2023ultra} identifies meta-topology types in KG structures, enabling zero-shot transfer through dataset-agnostic representations of entities and relations.
% However, ULTRA is 
Nevertheless, it remains limited to structural information and does not incorporate textual modalities, which are critical for contextual reasoning. Moreover, it focuses exclusively on in-KG reasoning tasks, neglecting out-of-KG tasks.

\paragraph{Text-aware Knowledge Graph Completion}
While earlier studies emphasized KG structures, recent work explores textual information for improved reasoning.
% KGBert transforms link prediction into a text classification problem using Transformer models \cite{yao2019kgbertbertknowledgegraph_kgbert}. 
% Methods like BertForLinkPrediction
BLP and StAR enhance representation learning by initializing embedding tables with language models (LMs) \cite{Daza_2021_blp, Wang_2021_star}. StATik \cite{markowitz-etal-2022-statik} combines LMs and graph neural networks (GNNs) by encoding node text with LMs and capturing structural information via message passing.

% These methods effectively integrate textual and structural modalities but rely on fine-tuning, constraining their generalization to unseen datasets or tasks \cite{galkin2023ultra}. 
Although these methods integrate textual and structural modalities effectively, their reliance on fine-tuning limits generalization to unseen datasets or tasks \cite{galkin2023ultra}.
Additionally, they remain limited to in-KG reasoning tasks and lack the flexibility to address out-of-KG tasks, such as Knowledge Graph Question Answering (KGQA), which demands broader integration of textual and structural information.

% The aforementioned studies have largely focused on the structural aspect of KGs. A complementary line of research explores how textual information within KGs can be leveraged to improve reasoning tasks. For instance, KGBert transforms link prediction tasks into text classification problems by utilizing Transformer models \cite{yao2019kgbertbertknowledgegraph_kgbert}. Similarly, BertForLink Prediction and StAR build upon this approach by integrating LMs to initialize embedding tables, thereby enhancing their representation capabilities \cite{Daza_2021_blp, Wang_2021_star}. Furthermore, StATik combines LMs with graph neural networks (GNNs) by encoding node text using an LM and then applying a message-passing neural network (MPNN) to capture structural information from neighbors \cite{markowitz-etal-2022-statik}.

% While these methods effectively integrate textual and structural modalities, they adhere to the "train-test" paradigm, where models are fine-tuned on labeled training data to perform inference on test data. This reliance on fine-tuning constrains their ability to generalize to unseen datasets or tasks, as demonstrated in \cite{galkin2023ultra}. Additionally, these models remain restricted to within-KG reasoning tasks and do not extend to beyond-KG reasoning tasks, such as KGQA, which require a broader integration of structural and textual modalities.

\paragraph{Knowledge Graph Question Answering}
KGQA represents a key out-of-KG reasoning task. It links topic entities in queries to detailed KG, improving answer accuracy through relational and contextual reasoning \cite{Wang_Kapanipathi_Musa_Yu_Talamadupula_Abdelaziz_Chang_Fokoue_Makni_Mattei_Witbrock_2019_qatower1}.

Early methods used dual-tower architectures combining graph- and textual features with minimal interaction between modalities \cite{yang-etal-2019-enhancing-pre_qatower2}. Later approaches trained LMs on KG data to extract implicit knowledge and generate effective subgraphs for QA \cite{mihaylov-frank-2018-knowledgeable_qatower3, lin-etal-2019-kagnet_qatower4, feng-etal-2020-scalable_qatower5, Lv_Guo_Xu_Tang_Duan_Gong_Shou_Jiang_Cao_Hu_2020_qatower6}.

Recent advancements include QA-GNN, which jointly updates LM and GNN layers through message passing \cite{yasunaga2021qagnn}, and GreaseLM, which enhances LM-GNN integration by aligning GNN and Transformer layers for comprehensive information fusion \cite{zhang2021greaselm}.

However, KGQA methods focus solely on out-of-KG reasoning tasks, while most KGC methods are confined to in-KG reasoning. This task-specific specialization highlights a key limitation: the lack of a unified framework capable of addressing both in-KG and out-of-KG reasoning effectively.

\vspace{-2mm}
\section{Task Definition}
A KG with textual information is defined as $\mathcal{G} = \{\mathcal{E}, \mathcal{R}, \mathcal{T}, \mathcal{D}\}$, where $\mathcal{E}$ and  $\mathcal{R}$  are  the set of entities and relations, $\mathcal{D}$ is the set of textual descriptions for entities and relations. The set of factual triples in the KG is denoted as $\mathcal{T} = \{(e_h, r, e_t)|$
$e_h,e_t \in \mathcal{E},r \in \mathcal{R} \}$, where $e_h,e_t\in\mathcal{E}$ and $r\in\mathcal{R}$. 

\paragraph{Inductive KGC.} KG Completion (KGC) task aims to predict the correct entity $e$ from the given KG $\mathcal{G}$ for query $(h,r,?)$ or $(?,r,t)$. In particular, inductive KGC tasks aim to train a score function based on the train KG $\mathcal{G}_{tr}=\{\mathcal{E}_{tr}, \mathcal{R}_{tr}, \mathcal{T}_{tr}, \mathcal{D}_{tr}\}$. Considering the different inductive settings of the test KG $\mathcal{G}_{te}=\{\mathcal{E}_{te}, \mathcal{R}_{te}, \mathcal{T}_{te}, \mathcal{D}_{te}\}$, we can categorize the evaluation into: \textbf{(1) KG containing only unseen entities}, which satisfies $\mathcal{E}_{tr} \neq \mathcal{E}_{te}$ and $\mathcal{R}_{tr} = \mathcal{R}_{te}$; \textbf{(2) KG containing both unseen entities and unseen relations,} which satisfies $\mathcal{E}_{tr} \neq \mathcal{E}_{te}$ and $\mathcal{R}_{tr} \neq \mathcal{R}_{te}$.
% The evaluation datasets are categorized into two types: (1) those containing only unseen entities and (2) those including both unseen entities and unseen relations. For datasets with only unseen entities, it is required to $\mathcal{E}_{train} \neq \mathcal{E}_{test}$ and $\mathcal{R}_{train} = \mathcal{R}_{test}$, whereas for the more challenging case of unseen entities and relations, it is necessary to $\mathcal{E}_{train} \neq \mathcal{E}_{test}$ and $\mathcal{R}_{train} \neq \mathcal{R}_{test}$.

\paragraph{KGQA.} 
% Knowledge Graph Question Answering (KGQA) is a reasoning task that leverages KGs as auxiliary knowledge bases to extract and reson. 
Given a query $question$ and several answer options $\mathcal{C}$, the KGQA task aims to retrieve subgraph from the KG $\mathcal{G}$ and predict the correct answer $a\in\mathcal{C}$. To maintain consistency with the KGC task format, we define query as $q=(question, REL\_the\_answer\_is,?)$, where 
% $question$ denotes the query question, 
$REL\_the\_answer\_is$ is an auxiliary relation specifically introduced to establish a connection between the query and its corresponding correct answer node. Additionally, a subgraph retrieved from the whole KG is represented as $\mathcal{G}_{sub} = \{ \mathcal{E}_{sub}, \mathcal{R}_{sub}, \mathcal{T}_{sub}, \mathcal{D}_{sub}\}$ with entities $\mathcal{E}_{sub} = \{ \mathcal{E}_{topic}, \mathcal{E}_{option}, \mathcal{E}_{other} \}$, where $\mathcal{E}_{topic}$ represents the entity mentioned in the question $q$, $\mathcal{E}_{option}$ represents the entity mentioned in the options, and $\mathcal{E}_{other}$ encompasses entities within the subgraph that do not carry particular contextual significance. The goal is to identify the correct answer option such that the triple 
$(question, REL\_the\_answer\_is,answer)$ is logically valid.

\section{Methodology}

\begin{figure*}[htbp]   %注意，这里设置是关键
% \centering\includegraphics[width=\linewidth,scale=1.00]{arch.png}
\centering\includegraphics[width=\linewidth,scale=1.00]{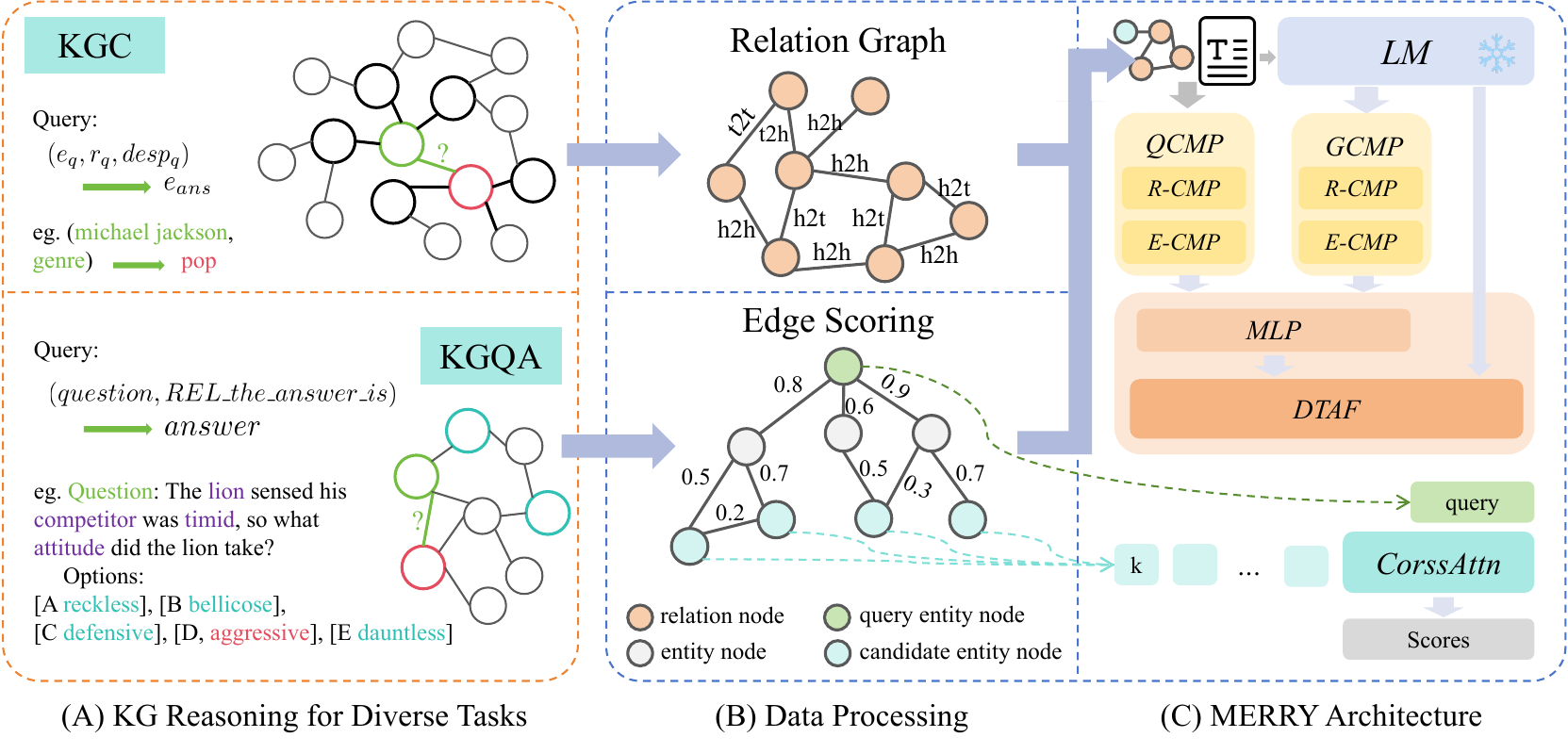}
	%[]里面的参数自己可根据需要调整
    \vspace{-6mm}
	\caption{
    % Overview of the General Knowledge Graph Reasoning Framework, MERRY. Given a query and KG, a relation graph is constructed using meta-topological relationships (e.g., h2h, h2t, etc.). A flexible edge scoring mechanism is employed, tailored to different downstream tasks. Within the encoder-decoder architecture, a cross-attention mechanism is utilized after multimodal information fusion to compute and predict the final answer.
    Overview of the MERRY Framework.
(A) All tasks, including KGC and KGQA, are unified under a standardized query representation.
(B) The data processing pipeline comprises two main components: (1) relation graph construction to model meta-relations, and (2) edge scoring to assign task-specific weights to edges.
(C) The MERRY architecture processes these graphs through QCMP, GCMP, and a multi-perspective dynamic fusion module. In the decoder, the query node is represented as the $Query$ embedding, while candidate nodes serve as $Key$ embeddings, outputting a probability distribution over all candidates.
    }
	\label{fig:arch}
    \vspace{-5mm}
\end{figure*}

A detailed breakdown of MERRY's components is presented in this section, as illustrated in Figure~\ref{fig:arch}. 
MERRY adopts an encoder-decoder architecture, and its processing can be formalized as follows:
\begin{align}
\vspace{-3mm}
    scores = \text{MERRY}(q,\mathcal{G},\mathcal{C})
    \vspace{-3mm}
\end{align}
where $q$ is the query, $\mathcal{G}$ is the graph containing relevant textual descriptions, and $\mathcal{C}$ are the candidates to be predicted. For KGC, $\mathcal{C}$ corresponds to candidate entities, while for KGQA, it includes all possible answer options. 
MERRY produces a probability distribution over the candidates, where higher scores reflect a higher likelihood of correctness.
% MERRY outputs a probability distribution over the candidates, with higher scores indicating a greater likelihood of correctness.

In the encoding phase, MERRY encodes the graph structure to derive its structural representation (Section \ref{sec:qcmp}) and explores strategies to effectively integrate textual and structural information (Section \ref{sec:gcmp}). A multi-perspective fusion module further enhances this process, enabling robust feature integration while preserving key textual semantics (Section \ref{sec:mpdf}). Additionally, we employ a flexible edge scoring mechanism to adapt to different tasks (Section~\ref{sec:edgescore}).

In the decoding phase, a flexible cross-attention decoder facilitates adaptation to diverse downstream tasks, including zero-shot KGC and KGQA.

% In the encoding phase, MERRY begins by encoding the graph structure to derive its structural representation. Subsequently, it explores efficient strategies for the seamless integration of textual and structural information. To further enhance this process, a multi-perspective fusion module is introduced, facilitating robust feature integration while preserving essential textual semantics. During the decoding phase, a flexible cross-attention decoder is employed, enabling smooth adaptation to diverse downstream tasks, including zero-shot KGC and KGQA.
% In the encoding phase, MERRY systematically explores strategies for effectively integrating textual and structural information (\ref{sec:cmp}). It also presents an efficient method for incorporating LM (\ref{sec:util-text}) and introduces the DTAF module (\ref{sec:DTAF}), which is specifically designed to retain the original modality-specific features. During the decoding phase, a flexible cross-attention decoder is utilized to ensure seamless adaptation to diverse downstream tasks, e.g., zero-shot KGC and KGQA.

\subsection{Conditional Message Passing}
\label{sec:cmp}
MERRY adopts Conditional Message Passing (CMP) as the basic GNN unit. Compared to traditional message-passing neural networks (MPNNs) like GCN\cite{kipf2017semisupervised_gcn}, GAT\cite{veličković2018graphattentionnetworks_gat}, and GraphSAGE\cite{10.5555/3294771.3294869_graphsage}, CMP explicitly conditions the representation of a target node $v$ on both a source node $u$ and a query relation $r_q$. For detailed architectural specifications of this conditioning mechanism, see \citet{huang2023a}. This process generates pairwise contextualized representations that dynamically adapt to the structural and semantic constraints imposed by $(u, r_q)$, enabling direct modeling of triple-level interactions\cite{zhu2021neural, zhang2022redgnn, galkin2023ultra}. Formally, the CMP process can be defined as:
% CMP provides the capability to model triple-level interactions by conditioning node and edge representations on a specific query \cite{zhu2021neural, zhang2022redgnn, galkin2023ultra}. The process of CMP can be formalized as follows:
\begin{align}
    &\mathbf{H}_{node} = \text{INIT}(q) \\
    &\mathbf{H}_{node} = \text{CMP}(\mathbf{H}_{node}, \mathbf{H}_{edge}, \mathcal{G})
\end{align}
where INIT is a conditional initialization function that initializes node representations conditioned on query $q$. It can be flexibly adapted for specific scenarios, as demonstrated in subsequent sections. $\mathbf{H}_{node}$ represents the node representations, $\mathbf{H}_{edge}$ is a learnable matrix for edge representations, and $\mathcal{G}$ denotes the graph structure. Detailed descriptions of the CMP calculations are provided in Appendix \ref{app:cmp}.
In the following sections, we develop two core modules for structural and textual encoding based on CMP unit.

\subsection{Query Conditional Structural Encoding}
\label{sec:qcmp}
To handle the scenario of unseen relationships in arbitrary KGs, we follow previous works~\cite{galkin2023ultra, chen2021topologyawarecorrelationsrelationsinductive_tact}, using the raw entity graph $\mathcal{G}$ and four fixed meta-relations $\mathcal{R}_{meta}=\{h2h, h2t, t2h, t2t\}$ to construct the corresponding relation graph. The relation graph is denoted as $\mathcal{G}_r=\{\mathcal{R}, \mathcal{R}_{meta}, \mathcal{T}_r\}$, where the nodes are relations derived from the entity graph $\mathcal{G}$, and the edges correspond to four types of meta-relations $\mathcal{R}_{meta}$. Details on the construction of triple sets $\mathcal{T}_r$ can be found in the Appendix \ref{app:rel_g}.

The introduction of the relation graph enables us to encode arbitrary structures. To achieve this, we propose the \textbf{QCMP} module, which applies CMP updates sequentially on the relation graph and the entity graph. This process yields query conditioned representations for both relations and entities. Given a query $q=(e_q, r_q, ?)$ and a KG $\mathcal{G} = \{\mathcal{E}, \mathcal{R}, \mathcal{T}, \mathcal{D}\}$, we first extract its relation graph $\mathcal{G}_r$ and then encode it as follows:
\vspace{-2mm}
\begin{align}    
    &\mathbf{r}_{r} = \begin{cases}
\mathbf{1}^d, \quad\mathrm{if}\quad r=r_q \\
\mathbf{0}^d, \quad \mathrm{otherwise} 
\end{cases},\text{for\;} r \in\mathcal{R} \\
% \mathbf{R}_{q\_init} &= \big[\mathbf{r}_1; \mathbf{r}_2; \dots; \mathbf{r}_n\big]   \\
% \mathbf{R}_q &= \text{CMP}(\mathbf{R}_{q\_init}, \mathbf{R}_{fund}, \mathcal{G}_r) \\ 
&\mathbf{R}_q = \text{CMP}\big(\mathbin{||}^{\mathcal{|\mathcal{R}|}}_{r=1}\mathbf{r}_r, \mathbf{R}_{meta}, \mathcal{G}_r\big)
\end{align}
% where \(\mathbf{R}_{q\_init}\) represents the initialized representations for all relations,
where $\mathbin{||}$ is the concatenation operation, \(\mathbf{R}_{meta}\in\mathbb{R}^{4\times d}\) is a learnable matrix corresponding to the four types of meta-relations, and \(\mathcal{G}_r\) is the relation graph constructed from \(\mathcal{G}\). The conditional initialization function assigns an all-ones embedding \(\mathbf{1}^d\) to the query relation \(r_q\), while all other relations are initialized with an all-zeros embedding \(\mathbf{0}^d\), where \(d\) is the dimension of embeddings. The final output \(\mathbf{R}_q\) represents the query conditioned relation embeddings. Subsequently, we update the entity graph with CMP module:
\begin{align}
    &\mathbf{h}_{e} = \begin{cases}
\mathbf{R}_q[r_q], \; \mathrm{if}\; e=e_q \\
\mathbf{0}^d, \quad\quad \mathrm{otherwise} 
\end{cases},\text{for}\; e \in\mathcal{E} \\
% \mathbf{H}_{q\_init} &= \big[\mathbf{h}_1; \mathbf{h}_2; \dots; \mathbf{h}_n\big]   \\
% \mathbf{H}_q &= \text{CMP}(\mathbf{H}_{q\_init}, \mathbf{R}_{q}, \mathcal{G})
&\mathbf{H}_q = \text{CMP}\big(\mathbin{||}^{\mathcal{|\mathcal{E}|}}_{e=1}\mathbf{h}_e, \mathbf{R}_{q}, \mathcal{G}\big)
\end{align}
where the embedding of $r_q$ is used as the initialization for $e_q$, while all other entities are initialized to all-zero embeddings. The final output $\mathbf{H}_q$ represents the query conditioned entity embeddings.

\subsection{Global Structural Semantic Encoding}
\label{sec:gcmp}
Textual information, as intrinsic node information, can be considered global information for the nodes. However, directly merging it with the structural modality information output by QCMP can lead to ineffective fusion due to the significant difference in their semantic spaces. Therefore, we propose the \textbf{GCMP} module to eliminate the semantic gap and achieve a more comprehensive modality fusion.

Specifically, we employ a Large Language Model (LLM) to encode textual information. However, since CMP requires features for all nodes in the graph as input, the substantial size of LLM weights can lead to an out-of-memory (OOM) risk. Therefore, we adopt a parameter-free strategy that extracts the representation of the last token from the LLM output to derive textual features for all nodes. The process of GCMP can be formalized as follows:
\begin{align}
&\mathbf{R}_{g}=\text{CMP}(\mathbf{1}^{|\mathcal{R}|\times d}, \mathbf{\hat{R}}_{meta},\mathcal{G}_r)\\
&\mathbf{H}_{g}=\text{CMP}(\bm{\mathcal{X}}_e,\mathbf{R}_{g}, \mathcal{G})
\end{align}
where $\mathbf{\hat{R}}_{meta}\in\mathbb{R}^{4\times d}$ represents a learnable matrix for meta-relations from textual perspective, $\bm{\mathcal{X}}_e$ represents the textual embeddings of all entities obtained via the parameter-free strategy. Specifically, each relation is initialized as an all-ones embedding, while the entity graph uses the textual embeddings $\bm{\mathcal{X}}_e$ as the initial representations.
By applying this sequential CMP update process, we generate the global semantic embeddings for relations $\mathbf{R}_{g}$ and entities $\mathbf{H}_{g}$.

\subsection{Multi-Perspective Dynamic Fusion}
\label{sec:mpdf}
\paragraph{Multi-Channel CMP Fusion}
As discussed earlier, MERRY encodes entities and relations from both query-specific and global perspectives through QCMP and GCMP, respectively. To integrate the outputs of these two CMP channels, we employ a multi-layer perceptron (MLP) for fusion:
\begin{align}
    &\mathbf{R}_{CMP} = \text{MLP}\big([\mathbf{R}_q|| \mathbf{R}_g]\big) \\
    &\mathbf{H}_{CMP} = \text{MLP}\big([\mathbf{H}_q|| \mathbf{H}_g]\big)
\end{align}
\paragraph{Dynamic Text-Adaptive Fusion}
Although multi-channel CMP fusion bridges structural and textual information, empirical observations indicate that tasks such as KGC and KGQA place differing levels of emphasis on textual features. To accommodate this variability and dynamically preserve task-specific textual information, we further propose a \underline{D}ynamic \underline{T}ext-\underline{A}daptive \underline{F}usion (\textbf{DTAF}) module.
%Encoding textual information is challenging due to its complexity and diversity. 
Specifically, we adopt a parameterized cross-attention mechanism to encode input textual descriptions $d\in\mathcal{D}$ into fixed-length embeddings:
\begin{align}
\bm{\mathcal{X}} &= \text{Attn}\big(\mathbf{Q}_{token}, \text{LM}(d), \text{LM}(d)\big)
\end{align}
where $\mathbf{Q}_{token} \in \mathbb{R}^{k\times d}$ represents trainable query parameters, $k$ is a tunable hyperparameter, and $\text{LM}(d)$ serves as both the Key and Value in the cross-attention mechanism. DTAF aggregates token-level information into meaningful representations $\bm{\mathcal{X}}$ while avoiding information loss.

Building on the textual embeddings, DTAF adaptively fuses textual and structural features using learnable weights $\alpha$ and $\beta$, balancing their contributions based on task requirements:
\begin{align}
    &\bm{\mathcal{X}}_r = \text{Attn}\big(\mathbf{Q}_{token}, \text{LM}(\mathcal{D}_{r}), \text{LM}(\mathcal{D}_{r})\big) \\
    &\bm{\mathcal{X}}_e = \text{Attn}\big(\mathbf{Q}_{token}, \text{LM}(\mathcal{D}_{e}), \text{LM}(\mathcal{D}_{e})\big) \\
    &\quad\mathbf{R}_f = \alpha * \bm{\mathcal{X}}_r + (1-\alpha) * \mathbf{R}_{CMP},\\
    &\quad\mathbf{H}_f = \beta * \bm{\mathcal{X}}_e + (1-\beta) * \mathbf{H}_{CMP},
\end{align}
where $\mathcal{D}_{r}$ and $\mathcal{D}_{e}$ are the textual descriptions of relations and entities, respectively. The outputs $\bm{\mathcal{X}}_r$ and $\bm{\mathcal{X}}_e$ represent the textual features of relations and entities, respectively. The fused embeddings $\mathbf{R}_f$ and $\mathbf{H}_f$ are unified representations that integrate three different perspectives.

\subsection{Query Conditional Edge Scoring}
\label{sec:edgescore}
Edge scores in MPNNs are crucial for model performance and vary significantly across tasks. To adapt to these differences, we design a flexible module tailored to task-specific requirements.

In KGC tasks, most methods focus on message passing and aggregation, often setting all edge scores to 1 \cite{veličković2018graphattentionnetworks_gat}. But in KGQA tasks, noisy paths in the retrieved subgraph necessitate more refined edge scoring. Compared to node relevance scores, edge scores capture richer interactions among the head entity, relation, and tail entity, offering a more accurate relevance measure for the query \cite{yasunaga2021qagnn}.
For each edge $(h,r,t)$ in the subgraph, its query relevance is calculated using a bilinear layer:
\begin{align}
\eta &= \text{Norm}\big([\mathbf{x}_h||\mathbf{x}_r|| \mathbf{x}_t]^\top \mathbf{W} \mathbf{x}_q\big),
\end{align}
where $\mathbf{W}\in\mathbb{R}^{3d\times d}$ is the bilinear coefficient, $\mathbf{x}_h$, $\mathbf{x}_r$, $\mathbf{x}_t$, $\mathbf{x}_q\in\bm{\mathcal{X}}$ represent the textual features of $(h,r,t)$ and the query $q$, obtained using the parameter-free method introduced in Section \ref{sec:gcmp}. The output $\eta \in \mathbb{{R}}^{2\times1}$ includes relevance and irrelevance scores, normalized with a Softmax function. The relevance score is then used in the update function of CMP. 

\subsection{Training Mechanism}
\label{sec:train}
\paragraph{Self-Supervised Pre-Training}
% Thanks to the inductive design, the framework can perform entity-agnostic and relation-agnostic encoding for arbitrary graphs, enabling pre-training on any graph or hybrid graph. The pre-training task uses self-supervised link prediction. Following standard practices in the literature \cite{sun2019rotateknowledgegraphembedding_rotate, zhu2021neural}, MERRY is trained by minimizing the binary cross-entropy loss over positive and negative samples:
The encoding process of MERRY is both entity-agnostic and relation-agnostic, making it suitable for inductive scenarios and allowing pre-training on arbitrary or hybrid KGs. The pre-training task employs self-supervised link prediction, with binary cross-entropy loss for positive and negative samples \cite{sun2019rotateknowledgegraphembedding_rotate, zhu2021neural}:
\begin{equation}
\mathcal{L}=-\log p(q,ans)-\sum_{i=1}^{n}\frac{1}{n}\log(1-p(q, neg\_ans)),
\end{equation}
where $q$ is the query prefix of the triple $(h, r, ? )$, and $ans$ is the tail entity $t$ that makes $(h,r,t)$ valid in the knowledge graphNegative samples are generated by randomly selecting tail entities. MERRY is pre-trained on multiple hybrid KG datasets, which equips it with generalizable transferability across diverse knowledge graphs.

\paragraph{Task Adaptation}
For the KGC task, the model is evaluated in a zero-shot setting without fine-tuning, using the same process as pre-training.

For the KGQA task, input questions are summarized as a combination of the query and the retrieved subgraph. The query is formalized as $q=(question,REL\_the\_answer\_is)$, where the candidates are the possible options. The goal is to select the correct answer such that $(question,REL\_the\_answer\_is, answer)$ forms a valid triple, with $REL\_the\_answer\_is$ is a newly introduced relation.

We adapt the data in three steps. First, a question-node is introduced to represent the input question, connected to all topic entities via a new relation. Its text description is the question itself. Additionally, each candidate option is represented by an answer-node, connected to the entities in the option via a special relation. Its text description is the original text of the option. Finally, we introduce a new relation, $REL\_the\_answer\_is$, which connects the question-node to the correct answer-node.

Since $REL\_the\_answer\_is$ lacks neighboring nodes in the relation graph, we adopt a few-shot approach. Using Sentence-BERT\cite{reimers2019sentencebertsentenceembeddingsusing} we compute sentence embeddings for each question and retrieve the top-K most similar questions based on cosine similarity. These few-shot examples are used to enrich the instances of $REL\_the\_answer\_is$.

With these modifications, MERRY can seamlessly transfer to perform the KGQA task.

\begin{table*}[htbp]
    \centering
    % \caption{Zero-shot performance on 25 KG inductive reasoning datasets. Best results among baselines and MERRY are \textbf{bolded}. ``(3g)'' means training with three KGs. The second-best results are \underline{underlined}.}
    
    % , and ``MERRY-X'' denotes the results of reasoning with different semantic features
    \resizebox{\textwidth}{!}{%
    \begin{tabular}{lcccccccccccccccc}
        \toprule
        \multirow{2}{*}{\textbf{Methods}} & \multicolumn{2}{c}{\textbf{IndE(WN)}} & \multicolumn{2}{c}{\textbf{IndE(FB)}} & \multicolumn{2}{c}{\textbf{IndE(NL)}} & \multicolumn{2}{c}{\textbf{IndER(FB)}} & \multicolumn{2}{c}{\textbf{IndER(WK)}} & \multicolumn{2}{c}{\textbf{IndER(NL)}} & \multicolumn{2}{c}{\textbf{Total AVG}} &
        \multirow{2}{*}{\textbf{SOTA Num}} \\
        \cmidrule(lr){2-3} \cmidrule(lr){4-5} \cmidrule(lr){6-7} \cmidrule(lr){8-9} \cmidrule(lr){10-11} \cmidrule(lr){12-13} \cmidrule(lr){14-15}
         & MRR & Hits@10 & MRR & Hits@10 & MRR & Hits@10 & MRR & Hits@10 & MRR & Hits@10 & MRR & Hits@10 & \textbf{MRR} & \textbf{Hits@10} \\
        \midrule
        % WN FB NL ICPL indER FB WK NL avg
        Supervised SOTA & 0.640 & 0.734 & 0.477 & 0.636  & 0.464 & 0.654 & 0.166 & 0.296 & 0.152 & 0.244 & 0.296 & 0.481 & 0.366 & 0.507 & - \\
        \midrule
        ULTRA(3g) & 0.517 & 0.678 & 0.486 & \underline{0.667} & \underline{0.561} & 0.742 & \textbf{0.386} & \textbf{0.599} & \underline{0.254} & 0.403 & 0.393 & 0.561 & 0.433 & 0.608 & 4 / 24 \\
        % ULTRA(3g) & 0.517 & 0.678 & 0.486 & 0.667 & 0.561 & 0.742 & 0.296 & 0.433 & 0.386 & 0.599 & 0.254 & 0.403 & 0.393 & 0.561 & 0.433 & 0.608 \\
        % ULTRA(4g) & 0.567 & 0.689 & 0.491 & 0.670 & 0.616 & 0.803 & - & - & 0.387 & 0.598 & 0.251 & 0.415 & 0.398 & 0.588 & 0.451 & 0.627 \\
        % ULTRA(50g) & 0.558 & 0.664 & \underline{0.493} & 0.664 & \textbf{0.590} & \textbf{0.777} & \underline{0.382} & 0.585 & 0.251 & 0.406 & 0.397 & 0.582 & \textbf{0.445} & 0.613 \\
        ProLINK & 0.553 & 0.690 & \textbf{0.494} & \textbf{0.684} & 0.546 & \underline{0.759} & 0.372 & 0.591 & 0.234 & 0.393 & \textbf{0.400} & \textbf{0.590} & 0.433 & 0.618 & \underline{8 / 24} \\
        \midrule
        MERRY & \textbf{0.563} & \textbf{0.709} & 0.486 & 0.662 & \textbf{0.567} & \textbf{0.767} & \underline{0.378} & \underline{0.592} & \textbf{0.282} & \textbf{0.443} & \underline{0.397} & \underline{0.586} & \textbf{0.445} & \textbf{0.626} & \textbf{ 12 / 24} \\
        MERRY$_{PNA}$ & \underline{0.559} & \underline{0.694} & 0.484 & 0.660 & 0.560 & 0.754 & 0.359 & 0.584 & 0.261 & \underline{0.426} & 0.384 & 0.569 & 0.435 & \underline{0.615} & - \\
        \bottomrule
    \end{tabular}%
    }
    \vspace{-2mm}
    \caption{Zero-shot and supervised SOTA performance on 24 KG inductive reasoning datasets. The best results across baselines, supervised methods, and MERRY are \textbf{bolded}.
    % ,while ``(3g)'' indicates training with three KGs. 
    The second-best results are \underline{underlined}. The \textbf{SOTA Num} column indicates the number of datasets where each method achieves SOTA performance.}
    \vspace{-6mm}
    \label{tab:kgc}
\end{table*}

\section{Experiments}
We evaluate MERRY on 28 datasets across two tasks: Inductive Knowledge Graph Completion (KGC) and Knowledge Graph Question Answering (KGQA). Our evaluation focuses on the following research questions:
\textbf{RQ1}: How effective is MERRY in reasoning for \textbf{in-KG} tasks under a zero-shot setting?
\textbf{RQ2}: Can MERRY effectively transfer and generalize to \textbf{out-of-KG} tasks?
\textbf{RQ3}: What is the impact of key components on the performance of MERRY?
\textbf{RQ4}: How do key hyperparameters affect the performance of MERRY?

\subsection{Datasets and Metrics}
\paragraph{Inductive KGC}
We perform zero-shot inductive KGC experiments on 27 datasets, categorized by entity and relation visibility: (1)
\textbf{Inductive Entity (e) Datasets (IndE)}: These datasets feature unseen entities in the test set, with fixed relations. This category includes 12 datasets from \cite{Teru2020InductiveRP_grail}: WN18RR (WN), FB15k-237 (FB), and NELL-995 (NL), each with four different versions.
(2) \textbf{Inductive Entity and Relation (e, r) Datasets (IndER)}: These datasets include unseen entities and relations in the test set. This category comprises 13 graphs from \cite{lee2023ingraminductiveknowledgegraph}: FB15k-237 (FB) and Wikidata68K (WK), each with four versions, and NELL-995 (NL), which has five versions.
We report Mean Reciprocal Rank (MRR) and Hits@10 results.

\paragraph{KGQA}
We use CommonsenseQA (CSQA) dataset \cite{talmor2019commonsenseqaquestionansweringchallenge_csqa}, which focuses on commonsense reasoning. It consists of 12,102 multiple-choice questions. We follow the in-house split method from \cite{lin-etal-2019-kagnet_qatower4} for experiments and compare our results with several baseline models.
We report Accuracy (Acc) on the CSQA dataset.

For detailed information on datasets and metric computation formulas, refer to Appendix \ref{app:datasets} and Appendix \ref{app:metric_compute}, respectively.

\subsection{Baselines}
\paragraph{Inductive KGC}
We compare MERRY against state-of-the-art supervised methods and recent KG foundation models, including ULTRA and ProLINK \cite{galkin2023ultra, wang-etal-2024-llm-prolink}, for zero-shot learning. Here, ULTRA(3g) refers to pre-training on three graphs.

\vspace{-2mm}
\paragraph{KGQA}
For KGQA, we use a fine-tuned standard LM as the baseline for models without external knowledge. Additionally, we evaluate several LM+KG-based methods, including RN \cite{santoro2017simpleneuralnetworkmodule_qa_rn}, RGCN \cite{schlichtkrull2017modelingrelationaldatagraph_rgcn}, GconAttn \cite{wang2018improvingnaturallanguageinference_gconattn}, KagNet \cite{lin-etal-2019-kagnet_qatower4}, MHGRN \cite{feng-etal-2020-scalable_qatower5}, QA-GNN \cite{yasunaga2021qagnn}, and GreaseLM \cite{zhang2021greaselm}. Among these, the best-performing models synchronize updates between the LM and GNN, enabling mutual interaction between textual and structural modalities.
% to improve feature representations.

\vspace{-2mm}
\subsection{Implementation \& Training details}
We pre-train MERRY on three hybrid knowledge graph datasets: WN18RR, CoDEx-Medium, and FB15k237, to capture diverse relational structures and sparsity patterns \cite{dettmers2018convolutional2dknowledgegraph_wn18rr, toutanova-chen-2015-observed_fb15k237, safavi-koutra-2020-codex}. Based on ULTRA, we set QCMP to a 6-layer CMP and GCMP to a 3-layer CMP, with each hidden layer having a dimension 64. 
To enhance convergence, we employ a two-stage training strategy: (1) QCMP weights from ULTRA are frozen, and other modules, particularly GCMP, are trained. (2) All components are unfrozen, allowing QCMP and other modules to converge jointly. During training, the LM backbone remains frozen.

For Inductive KGC, we evaluate the zero-shot capability of the pre-trained model directly on downstream datasets, using the Llama3 8B LM backbone \cite{grattafiori2024llama3herdmodels}.

For KGQA, due to the substantial gap between pre-training and the downstream task, we fine-tune the model with three few-shot examples before testing. 
% Given the nature of commonsense reasoning, which
Considering commonsense reasoning 
requires alignment with human cognitive preferences, 
we use the Llama3 8B Instruct backbone.

\subsection{Main Results (RQ1)}
We compare MERRY with baselines on 27 inductive link prediction KG datasets, categorized into 7 benchmarks based on data sources. For a fair comparison, datasets IndE (ILPC-small), IndE (ILPC-large), and IndER (NL-0) are excluded. Table \ref{tab:kgc} presents the average results across 6 benchmarks, 24 datasets. A full comparison of results across 27 datasets is provided in Appendix \ref{app:full_res}.

Four benchmarks, IndE(X) from \cite{Teru2020InductiveRP_grail}, contain unseen entities in the test graph. In contrast, the IndER (X) benchmark from \cite{lee2023ingraminductiveknowledgegraph} includes unseen entities and relations, making it significantly more challenging.
Among all dataset benchmarks, IndER (WK), IndE (NL), and IndER (NL) contain entities and relations unseen during pre-training, providing a strong evaluation of the model's zero-shot generalization capability. Table \ref{tab:kgc} shows that MERRY outperforms baselines.

Additionally, we compare MERRY with a parameter-free PNA method \cite{corso2020principalneighbourhoodaggregationgraph_pna}, used for encoding textual descriptions of entities and relations (\ref{sec:gcmp}). From the average results, while the $\text{MERRY}_{PNA}$ variant shows a slight decline in performance, it demonstrates that our design retains a certain level of robustness.

Overall, MERRY surpasses state-of-the-art supervised models and existing zero-shot transfer methods in total average metrics. While ULTRA and ProLINK excel on specific datasets, their performance is largely limited to datasets they were trained on.

\begin{table}
  \centering
  \resizebox{0.9\columnwidth}{!}{%
  \begin{tabular}{lcc}
    \toprule
    \textbf{Methods} & \textbf{IHdev-Acc.}(\%) & \textbf{IHtest-Acc.}(\%) \\
    \midrule
    RoBERTa-Large     & 73.1 & 68.7           \\
    LLaMA-3-8b-instruct & 72.9 & 71.9 \\
    RGCN & 72.7 & 68.4          \\
    GconAttn & 72.6 & 68.6 \\
    KagNet & 73.5 & 69.0 \\
    RN & 74.6 & 69.1 \\
    MHGRN & 74.5 & 71.1 \\
    QA-GNN & 76.5 & 73.4 \\
    GreaseLM & \underline{78.5} & \underline{74.2} \\
    \midrule
    MERRY & \textbf{78.6} & \textbf{74.9} \\
    \bottomrule
  \end{tabular}
  }
  % \caption{\textbf{Performance comparison on CommonsenseQA in-house split} (controlled experiments). As the official test is hidden, here we report the in-house Dev (IHdev) and Test (IHtest) accuracy, following the data split of \cite{lin-etal-2019-kagnet_qatower4}}
  \vspace{-3mm}
  \caption{Performance comparison on CommonsenseQA in-house split (controlled experiments).}
  \label{tab:kgqa}
  \vspace{-6mm}
\end{table}

\subsection{Generalization to KGQA (RQ2)}
Table \ref{tab:kgqa} compares MERRY with previous state-of-the-art methods on the CSQA dataset. MERRY achieves superior performance, surpassing all baselines and delivering the best overall results. Notably, compared to GreaseLM, which integrates GNN and LM layers through bidirectional interactions, MERRY performs comparably on the validation set but exceeds it on the test set. This demonstrates the effectiveness of our approach in integrating textual and structural modalities.

These results highlight the robustness of our multimodal fusion strategy and strong generalization capabilities. Additionally, in zero-shot inference using Llama3 8b Instruct, MERRY shows significant improvement, further validating its ability to incorporate structural information without compromising textual understanding.

\subsection{Ablation Studies (RQ3)}
\begin{figure}[htbp]   %注意，这里设置是关键
	\vspace{-3mm}
	\centering
	
	\includegraphics[width=\linewidth,scale=1.00]{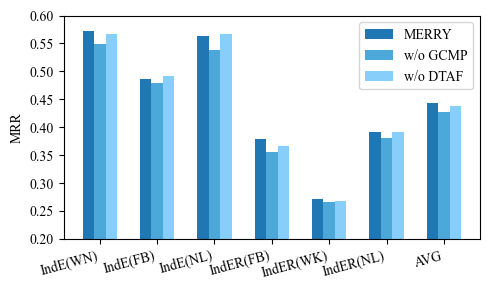}
	
	%[]里面的参数自己可根据需要调整
	
	% \caption{Ablation study results of different MERRY variants. "w/o GCMP" refers to the variant where the GCMP module is removed, and "w/o DTAF" denotes the variant that ignores the residual fusion of the original textual features.}
        \vspace{-3mm}
	\caption{Ablation study results.}
	
	\label{fig:ab-kgc}
	\vspace{-3mm}
\end{figure}

We conducted ablation experiments on multiple datasets, including IndE(X) and IndER(X), to evaluate the impact of two key components in our method for KGC. As shown in Figure \ref{fig:ab-kgc},
"w/o GCMP" indicates the removal of the GCMP module, where node text and structural features are instead concatenated and fused via an MLP. "w/o DTAF" refers to the model where DTAF is ignored, relying solely on CMP-based fusion for downstream predictions.

The results demonstrate a significant performance drop in the "w/o GCMP" variant, highlighting its critical role in bridging the gap between textual and structural modalities for better integration. In contrast, the "w/o DTAF" variant shows a slight performance decline, indicating that while original text features aid KGC, DTAF primarily enhances the understanding of structural information.

\begin{table}
  \centering
  \resizebox{\columnwidth}{!}{%
  \begin{tabular}{cc|cc}
    \toprule
    \textbf{Edge Scoring} & \textbf{DTAF} & \textbf{IHdev-Acc.}(\%) & \textbf{IHtest-Acc.}(\%) \\
    \midrule
    \checkmark & \checkmark & 78.6 & 74.9 \\
    & \checkmark & 77.7 & 75.0 \\
    & & 71.4 & 70.7 \\
    \bottomrule
  \end{tabular}
  }
    \caption{Ablation results of the edge scoring mechanism and DTAF module on the CSQA dataset.}
    \vspace{-5mm}
    \label{tab:ab-kgqa}
\end{table}

Similarly, we conducted ablation experiments on the CSQA dataset, as shown in Table \ref{tab:ab-kgqa}. An additional variant, "w/o Edge Scoring", sets all edge scores to 1, similar to the KGC tasks. The results indicate that DTAF significantly impacts KGQA performance, highlighting the importance of text feature understanding in these tasks and its role in preserving the LM's text processing capability. Moreover, ignoring edge scores results in a performance decline, underscoring the importance of edge weights in KGQA.

\subsection{Hyperparameter Sensitivity (RQ4)}
\begin{figure}[htbp]   %注意，这里设置是关键
	\vspace{-4mm}
	\centering
	
	\includegraphics[width=\linewidth,scale=1.00]{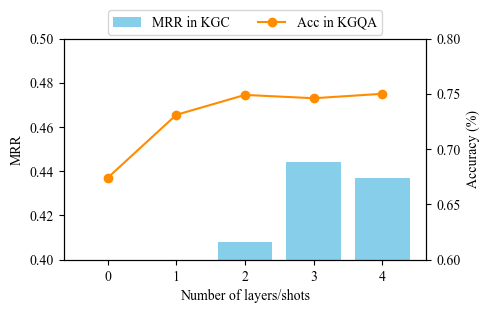}
	
	%[]里面的参数自己可根据需要调整
	\vspace{-3mm}
	\caption{Performance of different GCMP layers in KGC and different numbers of shots in KGQA.}
	
	\label{fig:key-p}
	\vspace{-3mm}
\end{figure}
We investigated the impact of GCMP layers on zero-shot KGC tasks and assessed the role of few-shot learning in KGQA. As illustrated in Figure \ref{fig:key-p}, using too few GCMP layers results in poor convergence, while excessive layers lead to feature smoothing. Aggregating information from up to three hops strikes an optimal balance, enabling effective performance.

For KGQA, the introduction of few-shot learning proves essential. As expected, zero-shot performance is initially poor. However, as the number of shots increases, performance stabilizes, demonstrating the model’s capacity to rapidly adapt and learn new relationships with minimal data.

\subsection{Computational Complexity and Scalability Analysis}
\label{sec:complexity}

To ensure practical applicability, we theoretically analyze MERRY's computational efficiency under two decoupled phases:

\begin{itemize}
    \item \textbf{Phase 1: LLM Text Encoding}  
    Complexity scales as $O(|V| \cdot T_{LLM})$, where $|V|$ is the node count and $T_{LLM}$ is the per-node encoding time. Our parameter-free feature extraction (Section \ref{sec:gcmp}) enables \textit{one-time offline preprocessing}, converting $T_{LLM}$ into a fixed cost during model deployment.

    \item \textbf{Phase 2: CMP Graph Updates}
Each iteration requires $O(|E|d + |V| d^2)$ operations, where $|E|$ denotes the number of edges and $d$ is the feature dimension. This complexity aligns with state-of-the-art GNNs like ULTRA~\cite{galkin2023ultra} and NBFNet~\cite{zhu2021neural}, while demonstrating significant advantages over classic inductive KGC approaches. Specifically, compared to GraIL's $O(|E|d ^2 +|V|d ^2)$  complexity for \textit{closed subgraph encoding} \cite{Teru2020InductiveRP_grail}, MERRY achieves a \textbf{$d$-fold reduction} in edge-related computation, making it particularly advantageous for graphs with large edge sets or high-dimensional features.
\end{itemize}

\noindent\textbf{Scalability Advantages:}
\noindent Based on the above time‐complexity analysis, MERRY demonstrates strong scalability on large‐scale graphs. By decoupling the LLM encoding phase, all node textual features can be precomputed offline at a cost of
$O\bigl(|V|\cdot T_{\mathrm{LLM}}\bigr)$
 and then stored and retrieved via a distributed system. Furthermore, the CMP graph-update complexity shows that, for a fixed hidden‐layer dimension $d$, MERRY’s online computation 
$O\bigl(|E|\,d + |V|\,d^2\bigr)$
is substantially lower than the $O\bigl(|E|\,d^2 + |V|\,d^2\bigr)$ required by classical approaches. Together, these results demonstrate that our framework achieves a favorable trade-off between performance and efficiency.

\section{Conclusion}
% \vspace{-1mm}
In this paper, we introduced MERRY, a general knowledge graph reasoning framework that bridges textual and structural modalities through multi-channel CMP encoding and multi-perspective dynamic fusion mechanisms. Additionally, we proposed a flexible edge scoring mechanism to adapt to diverse downstream tasks. Experiments across 28 datasets demonstrate MERRY’s strong generalization capabilities in in-KG tasks, such as zero-shot KGC, and its adaptability to out-of-KG tasks, such as KGQA, highlighting its potential as a unified framework for reasoning across in-KG and out-of-KG tasks.

\section*{Acknowledgment}
This work is founded by National Natural Science Foundation of China (NSFC62306276/NS-FCU23B2055/NSFCU19B2027), Zhejiang Provincial Natural Science Foundation of China (No. LQ23F020017), Yongjiang Talent Introduction Programme (2022A-238-G), and Fundamental Research Funds for the Central Universities (226-2023-00138). 

\section*{Limitations}
Here, we discuss three limitations of this work. First, through hyperparameter tuning experiments, it is evident that the CMP module's depth has limitations. A higher number of layers leads to feature smoothing, which is a challenge commonly faced by models incorporating GNN architectures. Second, we assumed that each entity and relation in the KG dataset has a corresponding textual description. However, our investigation discovered that some datasets need better maintenance, resulting in missing textual fields for certain entities. This issue of data completeness poses challenges for approaches that rely on language models. Finally, while LLM have demonstrated significant potential across various tasks, they face unique challenges in the in-KG task. Due to the size of the graph, encoding all nodes becomes particularly difficult, not only introducing substantial time and memory overhead during training but also consuming considerable storage space for offline feature storage. Efficiently leveraging LLMs in the in-KG tasks thus remains a crucial area for future exploration.

\bibliography{custom}

\appendix

\section{Details of CMP Updates}
\label{app:cmp}
Given a graph $\mathcal{G} = (\mathcal{E}, \mathcal{R}, \mathcal{T})$, where the feature of any entity $u$ $\mathbf{h}_u$ and the feature of any relation is denoted as $\mathbf{r}$, the update process for the $(t+1)$-th layer of CMP (Conditional Message Passing) is formalized as follows:
\begin{align}
\mathbf{m}_{u}^{t+1} &= \text{MSG}(\mathbf{h}_{w}^t,\mathbf{r}), w\in\mathcal{N}_r(u),\\
\mathbf{h}_{u}^{t+1} &=\text{UPDATE}\Big(\mathbf{h}_{u}^t,\text{AGG}(\mathbf{m}_{u}^{t+1})\Big)
\end{align}
where, we follow the settings of NBFNet, where the message function uses the parameter-free DistMult, the aggregation function employs summation, and UPDATE is implemented as a linear layer with LayerNorm.

When edge scores are introduced, the message function is adjusted to incorporate relevance scores. If the relevance score for any edge is denoted as $s$, the modified update equations become:
\begin{align} 
\mathbf{m}_{u}^{t+1} &= s \cdot \text{MSG}(\mathbf{h}_{w}^t,\mathbf{r}), w\in\mathcal{N}_r(u),\\
\mathbf{h}_{u}^{t+1} &=\text{UPDATE}\Big(\mathbf{h}_{u}^t,\text{AGG}(\mathbf{m}_{u}^{t+1})\Big)
\end{align}
where the edge score $s$ weights the message contribution from each neighbor, enhancing the model's ability to capture relevance-specific information in graph updates.

\section{Relation Graph Construction}
\label{app:rel_g}
Given a graph $\mathcal{G} = (\mathcal{E}, \mathcal{R}, \mathcal{T})$, we apply the lifting function $\mathcal{G}_r = \textsc{Lift}(\mathcal{G})$ to build a graph of relations $\mathcal{G}_r = (\mathcal{E}_r, \mathcal{R}_{meta}, \mathcal{T}_r)$ where each node is a distinct relation type in $\mathcal{G}$.
Triples $\mathcal{T}_r  \in (\mathcal{R} \times \mathcal{R}_{meta} \times \mathcal{R})$ in the relation graph $\mathcal{G}_r$ denote interactions between relations in the original graph $\mathcal{G}$, and we distinguish four such meta-relation interactions $\mathcal{R}_{\textit{meta}}$:  \emph{tail-to-head (t2h)} edges, \emph{head-to-head (h2h)} edges, \emph{head-to-tail (h2t)} edges, and \emph{tail-to-tail (t2t)} edges. Each of the four adjacency matrices can be efficiently obtained with one sparse matrix multiplication; for details, refer to \citet{galkin2023ultra}.

\section{Datasets}
\label{app:datasets}
\paragraph{Pre-Training}
Considering MERRY's effective generalization across datasets, we perform pre-training using a mix of the WN18RR, FB15k237, and CodexMedium datasets. Table \ref{tab:pre_dataset} presents the statistics of these three datasets, highlighting their data diversity.

\paragraph{Inductive KGC}
Our zero-shot Inductive KG Completion (KGC) experiments are conducted on 27 datasets. Among these, 12 datasets are derived from the GraIL framework \cite{Teru2020InductiveRP_grail}, which utilizes widely recognized KG benchmarks such as WN18RR \cite{dettmers2018convolutional2dknowledgegraph_wn18rr}, FB15k237 \cite{toutanova-chen-2015-observed_fb15k237}, and NELL-995 \cite{xiong2018deeppathreinforcementlearningmethod_nell955}, and 2 datasets are derived from the ILPC \cite{Galkin2022_ilpc2022}. These datasets are designed such that the training and testing graphs maintain consistent relation types.

Additionally, we incorporate 13 datasets from the InGram framework \cite{lee2023ingraminductiveknowledgegraph} to further assess inductive reasoning performance. These datasets are generated from three real-world knowledge graph benchmarks: FB15k237 \cite{toutanova-chen-2015-observed_fb15k237}, Wikidata68K \cite{10.1145/3579051.3579066_wikipedia}, and NELL-995 \cite{xiong2018deeppathreinforcementlearningmethod_nell955}. Each dataset is partitioned into subsets with varying proportions of novel relational triples, specifically 100\%, 75\%, 50\%, and 25\%, enabling evaluation under diverse inductive settings. Additionally, the NELL-995 also has a variant dataset with 0%.

While other KG datasets with textual descriptions exist, their limited accessibility precludes their inclusion in this study. Future research may focus on evaluating these datasets. Comprehensive structural statistics for the datasets employed in this work are presented in Table \ref{tab:kgc_datasets}.

\paragraph{KGQA}
In our KG question answering (KGQA) experiments, the CommonsenseQA dataset is used as a representative for this type of task \cite{talmor2019commonsenseqaquestionansweringchallenge_csqa}.
CSQA is a multiple-choice question-answering benchmark with five answer options per question, aimed at assessing reasoning based on commonsense knowledge. It includes a total of 12,102 questions. As the test set for CSQA is not openly accessible, evaluation can only be conducted biweekly through submissions to the official leaderboard.

For our primary experiments, we rely on the in-house (IH) data splits introduced by \cite{lin-etal-2019-kagnet_qatower4} for training and validation purposes. The performance of our final system is also evaluated on the official test set to provide a direct comparison with existing methods.

\begin{table}[ht]
\centering
\resizebox{\columnwidth}{!}{%
\begin{tabular}{lcccccc}
\hline
Dataset & $|\mathcal{E}_{tr}|$ & $|\mathcal{R}_{tr}|$ & \#Train & \#Validation & \#Test \\
\hline
WN18RR & 40.9k & 11 & 86.8k & 3.0k & 3.1k \\
FB15k-237 & 14.5k & 237 & 272.1k & 17.5k & 20.4k \\
CodexMedium & 17.0k & 51 & 185.5k & 10.3k & 10.3k \\
\hline
\end{tabular}}
    \caption{Statistics of pre-training KG datasets.}
\label{tab:pre_dataset}
\end{table}

\section{Metrics}
\label{app:metric_compute}
\paragraph{Mean Reciprocal Rank (MRR)}
The \textbf{Mean Reciprocal Rank (MRR)} evaluates the quality of the ranking in Knowledge Graph Completion (KGC) tasks. For a given query $q$, let the rank of the correct candidate be $r_q$. The reciprocal rank is defined as $\frac{1}{r_q}$. Averaging over all queries, MRR is calculated as:
\begin{align}
\text{MRR} &= \frac{1}{|\mathcal{Q}|} \sum_{q \in \mathcal{Q}} \frac{1}{r_q}
\end{align}
where $\mathcal{Q}$ represents the set of all queries. A higher MRR indicates better model performance in ranking the correct candidate higher in the prediction list.

\paragraph{Hits@10}
The \textbf{Hits@10} metric measures the proportion of queries for which the correct candidate is ranked within the top 10 predictions. For a given query $q$, let the rank of the correct candidate be $r_q$. Hits@10 is defined as:
\begin{equation}
\text{Hits@10} = \frac{1}{|\mathcal{Q}|} \sum_{q \in \mathcal{Q}} \mathbf{1}[r_q \leq 10],
\end{equation}
where $\mathbf{1}[\cdot]$ is an indicator function that equals 1 if the condition inside is true and 0 otherwise. A higher Hits@10 value reflects the model's ability to include the correct candidate within the top 10 ranked predictions.

\paragraph{Accuracy (Acc)}
The \textbf{Accuracy (Acc)} metric is used to evaluate performance on Knowledge Graph Question Answering (KGQA) tasks. For a dataset of queries, let $\mathbf{1}[q]$ indicate whether the predicted answer for question $q$ matches the ground truth. Accuracy is computed as:
\begin{equation}
\text{Acc} = \frac{1}{|\mathcal{Q}|} \sum_{q \in \mathcal{Q}} \mathbf{1}[q],
\end{equation}
where $\mathcal{Q}$ represents the set of all questions. A higher Accuracy score indicates the model's effectiveness in selecting the correct answer from the set of options.

\section{Full Results}
\label{app:full_res}
The full, per-dataset results of MRR and Hits@10 of the zero-shot inference of the pre-trained MERRY model, the pre-trained ULTRA model, and best reported supervised SOTA baselines are presented in Table \ref{tab:full_res}.

The detailed results from Table \ref{tab:kgc} are presented in Table \ref{tab:full_res}, which also includes the outcomes for two ILPC datasets ans IndER(NL-0) that are not covered in \cite{wang-etal-2024-llm-prolink}.

\begin{table*}[ht]
\centering
\resizebox{0.95\textwidth}{!}{%
\begin{tabular}{@{}cclcccccccccc@{}}
\toprule
\textbf{Group} & \textbf{Dataset} & \multicolumn{3}{c}{\textbf{Training Graph}} & \multicolumn{3}{c}{\textbf{Validation Graph}} & \multicolumn{3}{c}{\textbf{Test Graph}} & \textbf{SOTA} \\
\cmidrule(lr){3-5} \cmidrule(lr){6-8} \cmidrule(lr){9-11}
 & & \centering Entities & \centering Rels & \centering Triples & Entities & Rels & Triples & Entities & Rels & Triples &  \\ \midrule

IndE(WN) & WN:v1 & 2746 & 9 & 5410 & 2746 & 9 & 5410 & 922 & 9 & 1618 &\citet{zhu2021neural} \\
 & WN:v2 & 6954 & 10 & 15262 & 6954 & 10 & 15262 & 2757 & 10 & 4011 & \citet{zhu2021neural} \\
 & WN:v3 & 12078 & 11 & 25901 & 12078 & 11 & 25901 & 5084 & 11 & 6327 & \citet{zhu2021neural} \\
 & WN:v4 & 3861 & 9 & 7940 & 3861 & 9 & 7940 & 12334 & 9 & 7084 & \citet{zhu2023anetscalablepathbasedreasoning_a*net} \\ \midrule
 
IndE(FB) & FB:v1 & 1594 & 180 & 4245 & 1594 & 180 & 4245 & 1093 & 180 & 1993 & \citet{zhu2023anetscalablepathbasedreasoning_a*net} \\
 & FB:v2 & 2608 & 200 & 9739 & 2608 & 200 & 9739 & 1660 & 200 & 4145 & \citet{zhu2021neural} \\
 & FB:v3 & 3668 & 215 & 17986 & 3668 & 215 & 17986 & 2501 & 215 & 7406 & \citet{zhu2021neural} \\
 & FB:v4 & 4707 & 219 & 27203 & 4707 & 219 & 27203 & 3352 & 219 & 11714 & \citet{zhu2023anetscalablepathbasedreasoning_a*net} \\ \midrule

IndE(NL) & NL:v1 & 3103 & 14 & 4687 & 3103 & 14 & 4687 & 833 & 14 & 833 & \citet{zhang2022redgnn} \\
 & NL:v2 & 2564 & 88 & 8219 & 2564 & 88 & 8219 & 2086 & 88 & 4586 & \citet{zhang2022redgnn} \\
 & NL:v3 & 4647 & 142 & 16393 & 4647 & 142 & 16393 & 3566 & 142 & 8048 & \citet{zhang2022redgnn} \\
 & NL:v4 & 2092 & 76 & 7546 & 2092 & 76 & 7546 & 2795 & 76 & 7073 & \citet{zhang2022redgnn} \\ \midrule
 
IndE(ILPC) & ILPC:small & 10230 & 48 & 78616 & 6653 & 48 & 2908 & 6653 & 48 & 2902 & \citet{galkin2022nodepiececompositionalparameterefficientrepresentations_nodepiece} \\
 & ILPC:large & 46626 & 65 & 202446 & 29246 & 65 & 10179 & 29246 & 65 & 10184 & \citet{galkin2022nodepiececompositionalparameterefficientrepresentations_nodepiece} \\ \midrule

IndER(FB) & FB-25 & 5190 & 163 & 91571 & 4097 & 216 & 17147 & 5716 & 4097 & 17147 & \citet{lee2023ingraminductiveknowledgegraph} \\
 & FB-50 & 5190 & 153 & 85375 & 4445 & 205 & 11636 & 3879 & 4445 & 11636 & \citet{lee2023ingraminductiveknowledgegraph} \\
 & FB-75 & 4659 & 134 & 62809 & 2792 & 186 & 9316 & 3106 & 2792 & 9316 & \citet{lee2023ingraminductiveknowledgegraph} \\
 & FB-100 & 4659 & 134 & 62809 & 2624 & 77 & 6987 & 2329 & 2624 & 6987 & \citet{lee2023ingraminductiveknowledgegraph} \\ \midrule
IndER(WK) & WK-25 & 12659 & 47 & 41873 & 3228 & 74 & 3391 & 1310 & 3228 & 3391 & \citet{lee2023ingraminductiveknowledgegraph} \\
 & WK-50 & 12022 & 72 & 82481 & 9328 & 93 & 9672 & 3224 & 9328 & 9672 & \citet{lee2023ingraminductiveknowledgegraph} \\
 & WK-75 & 6853 & 52 & 28741 & 2722 & 65 & 3430 & 1143 & 2722 & 3430 & \citet{lee2023ingraminductiveknowledgegraph} \\
 & WK-100 & 9784 & 67 & 49875 & 12136 & 97 & 13487 & 4496 & 12136 & 13487 & \citet{lee2023ingraminductiveknowledgegraph} \\ \midrule
IndER(NL) & NL-0 & 1814 & 134 & 7796 & 2026 & 112 & 2287 & 2026 & 112 & 2287 & \citet{lee2023ingraminductiveknowledgegraph} \\
& NL-25 & 4396 & 106 & 17578 & 2230 & 146 & 2230 & 743 & 2230 & 2230 & \citet{lee2023ingraminductiveknowledgegraph} \\
 & NL-50 & 4396 & 106 & 17578 & 2335 & 119 & 2576 & 859 & 2335 & 2576 & \citet{lee2023ingraminductiveknowledgegraph} \\
 & NL-75 & 2607 & 96 & 11058 & 1578 & 116 & 1818 & 607 & 1606 & 1818 & \citet{lee2023ingraminductiveknowledgegraph} \\
 & NL-100 & 1258 & 55 & 7832 & 1709 & 53 & 2378 & 793 & 1709 & 2378 & \citet{lee2023ingraminductiveknowledgegraph} \\ \bottomrule
\end{tabular}%
}
\caption{Inductive KG datasets used in the experiments. "Triples" refers to the number of edges in the graph used for training, validation, or testing. "Valid" and "Test" refer to the triples that need to be predicted in the validation and test sets, respectively, within the corresponding graphs.}
\label{tab:kgc_datasets}
\end{table*}

\begin{table*}[ht]
\centering
\resizebox{0.65\textwidth}{!}{%

\begin{tabular}{@{}cclccccc@{}}
\toprule
\textbf{Group} & \textbf{Dataset} & \multicolumn{2}{c}{\textbf{ Supervised SOTA }} & \multicolumn{2}{c}{\textbf{ULTRA(3g)}} & \multicolumn{2}{c}{\textbf{MERRY}} \\
% & \textbf{SOTA} \\
\cmidrule(lr){3-4} \cmidrule(lr){5-6} \cmidrule(lr){7-8}
 & & MRR &  Hits@10 &  MRR & Hits@10 & MRR & Hits@10   \\ \midrule

IndE(WN) & WN:v1 & 0.741 & 0.826 & 0.593 & 0.779 & 0.635 & 0.795   \\
 & WN:v2 & 0.704 & 0.798 & 0.620 & 0.752 & 0.654 & 0.783     \\
 & WN:v3 & 0.452 & 0.568 & 0.371 & 0.494 & 0.397 & 0.526     \\
 & WN:v4 & 0.661 & 0.743 & 0.484 & 0.687 & 0.562 & 0.710     \\ \midrule
 
IndE(FB) & FB:v1 & 0.457 & 0.589 & 0.486 & 0.657 & 0.478 & 0.628  \\
 & FB:v2 & 0.51 & 0.672 & 0.501 & 0.694 & 0.503 & 0.694     \\
 & FB:v3 & 0.476 & 0.637 & 0.482 & 0.644 & 0.478 & 0.636     \\
 & FB:v4 & 0.466 & 0.645 & 0.477 & 0.671 & 0.484 & 0.688      \\ \midrule

IndE(NL) & NL:v1 & 0.637 & 0.866 & 0.716 & 0.861 & 0.643 & 0.892  \\
 & NL:v2 & 0.419 & 0.601 & 0.525 & 0.719 & 0.558 & 0.753      \\
 & NL:v3 & 0.436 & 0.594 & 0.511 & 0.687 & 0.564 & 0.730      \\
 & NL:v4 & 0.363 & 0.556 & 0.490 & 0.701 & 0.498 & 0.691      \\ \midrule
 
IndE(ILPC) & ILPC:small & 0.130 & 0.251 & 0.302 & 0.443 & 0.335 & 0.472      \\
 & ILPC:large & 0.070 & 0.146 & 0.290 & 0.424 & 0.302 & 0.437      \\ \midrule

IndER(FB) & FB-25 & 0.133 & 0.271 & 0.383 & 0.633 & 0.363 & 0.616  \\
 & FB-50 & 0.117 & 0.218 & 0.330 & 0.536 & 0.330 & 0.540    \\
 & FB-75 & 0.189 & 0.325 & 0.391 & 0.594 & 0.377 & 0.574     \\
 & FB-100 & 0.223 & 0.371 & 0.438 & 0.631 & 0.443 & 0.638      \\ \midrule
IndER(WK) & WK-25 & 0.186 & 0.309 & 0.307 & 0.507 & 0.293 & 0.487  \\
 & WK-50 & 0.068 & 0.135 & 0.158 & 0.296 & 0.216 & 0.402    \\
 & WK-75 & 0.247 & 0.362 & 0.373 & 0.519 & 0.401 & 0.531      \\
 & WK-100 & 0.107 & 0.169 & 0.178 & 0.289 & 0.220 & 0.360      \\ \midrule
IndER(NL) & NL-0 & 0.269 &0.431  & 0.342 & 0.523 & 0.351 & 0.536      \\
& NL-25 & 0.334 & 0.501 & 0.387 & 0.538 & 0.406 & 0.601      \\
 & NL-50 & 0.281 & 0.453 & 0.398 & 0.549 & 0.376 & 0.530      \\
 & NL-75 & 0.261 & 0.464 & 0.348 & 0.527 & 0.344 & 0.550      \\
 & NL-100 & 0.309 & 0.506 & 0.442 & 0.631 & 0.462 & 0.666      \\ \bottomrule
\end{tabular}%
}
\caption{The full results (MRR and Hits@10) of MERRY, ULTRA, and the best-reported Supervised SOTA are presented across 27 datasets, highlighting their performance under both zero-shot inference and fine-tuning scenarios.}
\label{tab:full_res}
\end{table*}

\end{document}